\documentclass[letterpaper, 10 pt, conference]{ieeeconf}  
\usepackage{algorithm}
\usepackage{indentfirst}
\usepackage{booktabs}
\usepackage{makecell}
\usepackage{multirow}
\usepackage{bbding}
\usepackage{graphicx}
\usepackage{amssymb}
\usepackage{tabularx}
\usepackage{algorithm}
\usepackage{algpseudocode}
\usepackage{amsmath}
\usepackage{listings}
\usepackage{longtable}
\usepackage{siunitx}
\usepackage{hyperref}
\newcolumntype{Y}{>{\centering\arraybackslash}X}

\IEEEoverridecommandlockouts                              

\title{EIMC:Efficient Instance-aware Multi-modal Collaborative Perception
}

\author{Kang Yang$^{1}$, Peng Wang$^{1}$, Lantao Li$^{2}$, Tianci Bu$^{3}$, Chen Sun$^{2}$, Deying Li$^{1}$, Yongcai Wang$^{1*}$%
\thanks{$^{*}$Corresponding author: Yongcai Wang}%
\thanks{$^{1}$School of Information, Renmin University of China, Beijing, China, 100872}%
\thanks{$^{2}$Sony Research and Development Center China, Beijing, China}%
\thanks{$^{3}$National University of Defense Technology, Hunan, China, 410073}%
}

\begin{document}

\maketitle
\thispagestyle{empty}
\pagestyle{empty}

\begin{abstract}

Multi-modal collaborative perception calls for great attention to enhancing the safety of autonomous driving. 
However,
current multi-modal approaches remain a ``local fusion $\to$ communication” sequence, which fuses multi-modal data locally and needs high bandwidth to transmit an individual's feature data before collaborative fusion. 
EIMC innovatively proposes an early collaborative paradigm. It injects lightweight collaborative voxels, transmitted by neighbor agents, into the ego’s local modality-fusion step, yielding compact yet informative 3D collaborative priors that tighten cross-modal alignment.
Next, a heatmap-driven consensus protocol identifies exactly where cooperation is needed by computing per-pixel confidence heatmaps. Only the Top-$K$ instance vectors located in these low-confidence, high-discrepancy regions are queried from peers, then fused via cross-attention for completion. Afterwards, we apply a refinement fusion that involves collecting the top-$K$ most confident instances from each agent and enhancing their features using self-attention. The above instance-centric messaging reduces redundancy while guaranteeing that critical occluded objects are recovered.
Evaluated on OPV2V and DAIR-V2X, EIMC attains 73.01\% AP@0.5 while reducing byte bandwidth usage by 87.98\% compared with the best published multi-modal collaborative detector. Code publicly released at \url{https://github.com/sidiangongyuan/EIMC}.

\end{abstract}

\section{INTRODUCTION}

Precise 3D scene perception is critical for intelligent agents in autonomous driving, drone technology, and robotics, enabling accurate situational awareness and real-time decision-making vital for operational safety. However, individual agent perceptual capabilities are inherently limited by restricted sensing ranges and occlusion effects, impeding comprehensive scene understanding. For instance, a single vehicle's sensors in autonomous driving may fail to detect occluded objects or long-range threats. These limitations necessitate a more holistic approach that aggregates and leverages information across multiple vantage points. Consequently, researchers have turned to collaborative perception, allowing agents to share complementary information for a more accurate and robust perception of their surroundings , thereby addressing occlusion and extending effective sensing ranges \cite{cooperative-development1,cooperative-development2}.

\begin{figure}[t]
    \centering
    \includegraphics[width=\linewidth]{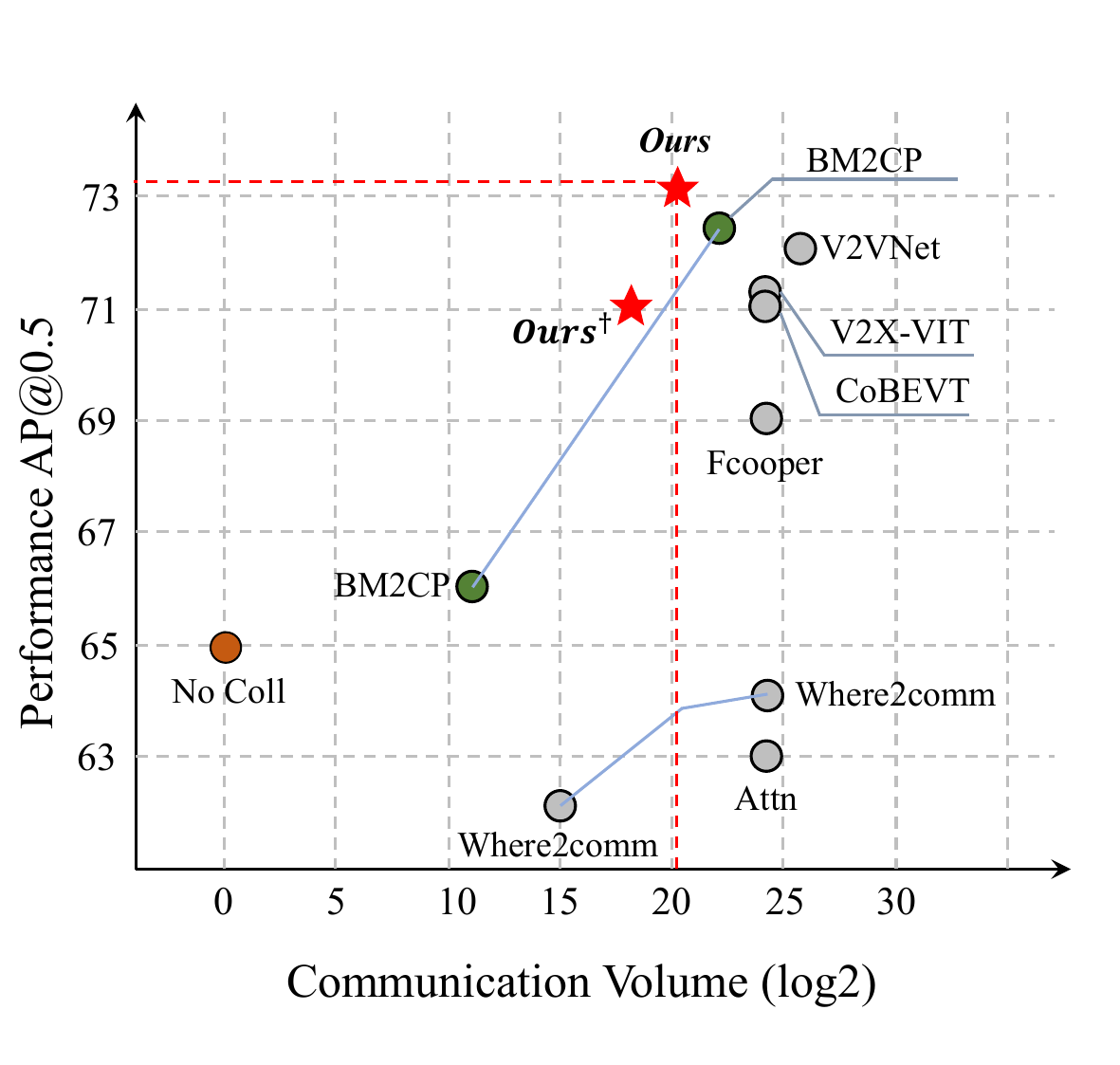}
    \caption{Compared with other intermediate fusion methods, EIMC achieves lower communication volume while still attaining the best performance. The $\text{Ours}^\dagger$ variant represents the version without the Mix-Voxel module. BM2CP \cite{bm2cp} is the multimodal-based collaborative perception work.}
    \label{introduction_fig}
\end{figure}

Current collaborative perception methodologies confront two critical challenges: 1) The multi-modal frameworks that achieved remarkable success in single-agent systems have not been sufficiently explored or effectively adapted for collaborative perception tasks; 2) The inherent communication overhead introduced by collaborative perception poses significant challenges to meet the stringent real-time requirements of autonomous driving systems. Early approaches \cite{v2vnet,v2x-vit,OPV2V,disconet} prioritized performance improvements while neglecting communication bandwidth constraints, resulting in limited practicality. Subsequent methods \cite{where2comm,TransIFF,Who2comm} attempted to balance communication efficiency with performance metrics, yet struggle to simultaneously maintain high performance and reduce communication demands. 
Such methods, for instance, have explored utilizing confidence maps for information filtering, but frequently encounter difficulties in accurately identifying all regions containing potential objects, and often still rely on transmitting dense BEV features, thereby hindering effective communication cost reduction. 
Recently, BM2CP \cite{bm2cp} pioneered the exploration of multi-modal collaborative perception frameworks, demonstrating the practicality of applying such approaches in this field. However, there is still room for optimization in multimodal fusion and balancing communication overhead and system performance. 
Overall, the most critical problem is how to optimally select essential perceptual information and enhance cross-agent knowledge transfer while ensuring communication efficiency.

In this work, we propose EIMC, a multimodal collaborative perception framework that jointly optimizes fusion accuracy and communication efficiency. EIMC advances collaboration by first injecting lightweight cross-agent voxels into the ego's modality-fusion step, creating compact, discriminative 3D collaborative priors to enhance cross-modal alignment.  Building upon this, a heatmap-driven consensus protocol precisely identifies cooperation needs, allowing only critical instance vectors from low-confidence, high-discrepancy regions to be queried from peers and fused for completion and refinement. This instance-centric messaging reduces redundancy while recovering occluded objects.

Specifically, the Modality Fusion stage employs the Mix-Voxel module to create a collaborative geometric prior from aggregated LiDAR voxels, which an occupancy head uses to re-weight the ego camera voxel before BEV feature collapse. The Heterogeneous Modality Fusion module then aligns the resulting camera and LiDAR BEV features in latent space. In the Collaboration stage, the Instance Completion module locates uncertain regions via heatmap discrepancies to retrieve complementary instance vectors. Instance Refinement further refines these vectors through self-attention and integrates them into the BEV representation via cross-attention. Furthermore, a multi-scale approach bolsters cooperative perception for diverse scenarios. As illustrated in Fig. \ref{introduction_fig}, EIMC achieves robust and efficient multimodal cooperative perception, effectively balancing performance and communication overhead.

To summarize, our contributions are:
\begin{itemize} 
    \item EIMC offers an efficient yet robust multimodal-based collaborative perception framework, enhancing 3D perception performance while reducing communication overhead.
    \item The Mix-Voxel and HMF modules construct compact and expressive scene-level representations, addressing modality gaps and alignment issues. The Instance Completion and Refinement modules focus on essential instance-level message transmission and fusion, yielding more robust and holistic scene understanding.

    \item Extensive experiments on 3D detection benchmarks demonstrate that EIMC consistently outperforms existing methods in accuracy, communication efficiency, and robustness.

\end{itemize}

\section{Related Work}
\subsection{Collaborative perception}
Collaborative perception aims to enhance agents’ perceptual capabilities by sharing information across a communication network. Mainstream research in this domain typically utilized single-modality sensors, such as LiDAR or cameras, for 3D object detection. These methods can be broadly categorized into early, intermediate, and late fusion approaches. 
Early fusion usually involves transmitting raw sensor data \cite{Cooper1,Cooper2}, but requires substantial bandwidth, while late fusion, which shares network outputs, often fails to deliver optimal performance.
In contrast, recent studies have turned to intermediate fusion techniques, seeking a better balance between efficiency and accuracy \cite{v2vnet,disconet,where2comm,v2x-vit,CoBEVT,TransIFF,quest, codrma}. For instance, V2VNet \cite{v2vnet} employs a spatially-aware graph neural network (GNN) to aggregate features across multiple agents, and AttnFuse \cite{OPV2V} is the first to introduce an attention mechanism for modeling multi-agent interactions. Moreover, V2X-VIT \cite{v2x-vit} adapts vision transformers (ViTs) \cite{vit1} for vehicle-to-everything (V2X) communication using heterogeneous self-attention, while Where2comm \cite{where2comm} and CoSDH \cite{cosdh} focuses on determining the optimal fusion points to minimize communication bandwidth. Additionally, DiscoNet \cite{disconet} leverages knowledge distillation to combine the advantages of both early and intermediate fusion methods, and CoBEVT \cite{CoBEVT} presents the first generic collaborative perception framework for multi-camera-based cooperative BEV semantic segmentation. BM2CP \cite{bm2cp} pioneers the exploration of multimodal cooperative perception tasks.
Building upon these foundations, instance-aware methodologies have emerged through works like TransIFF's \cite{TransIFF, ACCO} transformer-based feature fusion and QUEST's \cite{quest} interpretable query cooperation. Beyond core perception tasks, subsequent research addresses practical deployment challenges including pose errors \cite{localizationproblem}, latency constraints \cite{latency1,latency2}, and multi-agent alignment \cite{Coalign,ROCO}. Recent studies, such as HEAL \cite{HEAL}, HM-VIT \cite{HM-VIT}, CodeFilling \cite{CodeFilling} and STAMP \cite{stamp}, focus on ensuring compatibility across heterogeneous agents.

\subsection{Multi-modal 3D Detection}
LiDAR and image data provide complementary information, enriching 3D scene understanding by combining precise geometric details with rich semantic cues. The integration of these modalities for 3D detection is an active area of research \cite{multimodal-task-1,multimodal-task-2}. One line of work is early fusion, exemplified by methods such as PointPainting and PointAugmenting \cite{pointpainting,pointaugmenting}, which augment LiDAR point clouds with features from camera images. Beyond early fusion, modern multimodal fusion strategies can be broadly divided into two categories. 
The first category uses a shared bird’s-eye-view (BEV) representation to fuse dense features from both LiDAR and camera modalities \cite{BEVFusion,BEVfusion-pku,deepinteraction,gafusion,msmdfusion}. For example, BEVFusion \cite{BEVFusion,BEVfusion-pku} leverages the Lift-Splat-Shoot (LSS) backbone \cite{LSS} to project image features into the BEV space and then concatenates them with LiDAR features, while DeepInteraction \cite{deepinteraction} keeps each modality’s features separate to enable efficient cross-modal interaction. 
The second category relies on sparse object queries to implicitly integrate features from both modalities \cite{UVTR,sparsefusion,futr3d,transfusion,focalformer3d}. TransFusion \cite{transfusion} illustrates this approach by introducing a query-based fusion strategy with a soft-association mechanism to handle poor image conditions, while Futr3D \cite{futr3d} presents a unified query-driven fusion framework. 
In this paper, we present EIMC, a novel and highly robust and efficient framework for multimodal collaborative perception that not only minimizes communication overhead but also sets a new benchmark in performance.

\section{Method}
\begin{figure*}[htbp]
    \centering
    \includegraphics[width=0.9\linewidth]{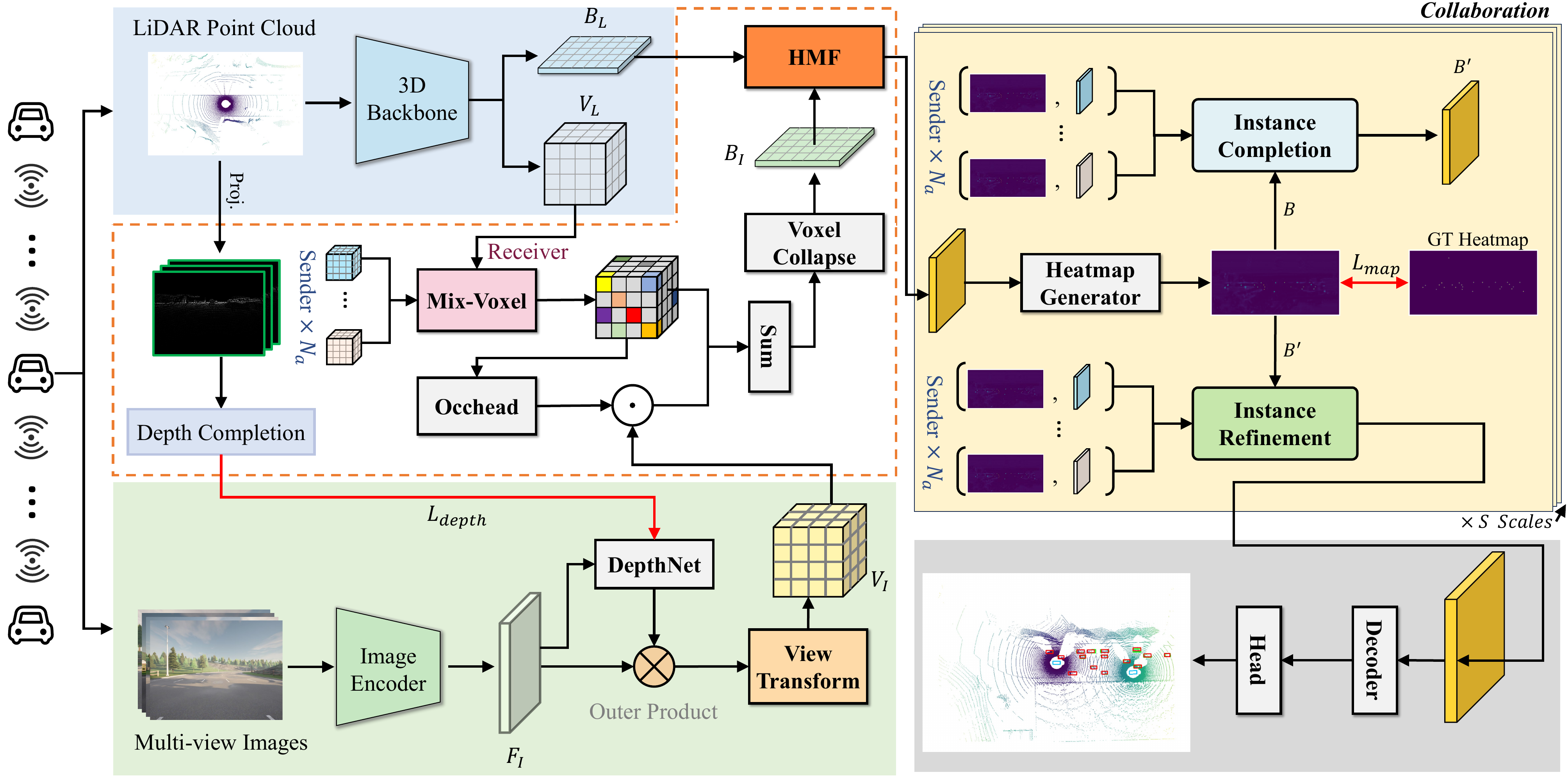}
    \caption{Framework. Given LiDAR and camera inputs, our method first extracts heterogeneous features through dedicated modality-specific encoders. The Mix-Voxel (MV) module leverages lightweight voxel transmission as priors to build the collaborative voxel and then constructs occupancy-guided voxel-based image representations, which are compressed into BEV features and fused with LiDAR BEV features through Heterogeneous Modality Fusion (HMF). Instance Completion (IC) and Instance Refinement (IR) modules subsequently propagate instance-level messages identified from heatmap priors. The collaboration employs multi-scale feature for final detection, with predicted and ground truth bounding boxes visualized as green and red boxes respectively.}
    \label{Framework}
\end{figure*}
\subsection{Problem Formulation}
The architecture of the EIMC is depicted in Fig. \ref{Framework}. In this scenario, we consider $N$ agents, and a communication graph $G$ models the interactions between all agents as vertices. Let $\mathbf{I}_n$ represent the RGB image collected by the camera, which may be captured by surrounding cameras, and let $\mathbf{L}_n$ denote the point cloud collected by the LiDAR of the $n$-th agent. Additionally, let $\mathcal{M}$ represent the message sent from neighboring agents to the ego agent $\mathcal{E}$, and $\mathbf{Y}_\mathcal{E}$ represent the perception supervision for the ego agent. The objective of collaborative perception is to achieve the maximized perception performance of all agents while hoping the communication cost is limited, that is:
\begin{equation}
\begin{gathered}
\arg\max_{\theta,\mathcal{M}} h\left( \Phi_\theta \left( \mathcal{M}_\mathcal{E}, \{\mathcal{M}_{n\to \mathcal{E}}\}_{n=1}^{N \neq \mathcal{E}} \right), \mathbf{Y}_\mathcal{E} \right), \\
\text{s.t. } \sum_{i=1}^N |\mathcal{M}_{i\to j}| \leq B,
\end{gathered}
\end{equation}
where $h(\cdot,\cdot)$ is the perception evaluation metric, $\Phi_\theta$ is the perception network with trainable parameter $\theta$. 
The process of our framework can be divided into two stages:
\begin{equation}
\left\{
\begin{array}{l}
\mathbf{B}_{\text{MF}}^n = f_\text{MF}(f_\text{enc}(\mathbf{I}_n, \mathbf{L}_n)), n=1,\cdots,N \quad \text{(stage-1)} \\[8pt]
\hat{\mathbf{Y}}_\mathcal{E} = f_{\text{dec}}\left( f_{\text{Col}}\left( \mathbf{B}_\mathcal{E}, \left\{ \mathcal{M}_{n \to \mathcal{E}} \right\}_{n \ne \mathcal{E}}^{N} \right) \right) \text{(stage-2)}
\end{array}
\right.
\end{equation}
Here, $f_\text{enc}$, $f_\text{MF}$, $f_{\text{dec}}$, and $f_{\text{Col}}$ denote the encoder, Modality Fusion stage, decoder, and collaboration stage, respectively. $\mathbf{B}_{\text{MF}}$ is the result of the modality fusion process, and $\hat{\mathbf{Y}}_\mathcal{E}$ represents the detection results.

\subsection{Modality Encoding}
\subsubsection{LiDAR branch.} Given the point clouds $\{\mathbf{P}_n\}_{n=1}^N$, we follow the mainstream approaches \cite{where2comm, v2vnet, OPV2V} and apply $f_\text{point}(\cdot)$ using PointPillars and VoxelNet. Formally, the LiDAR BEV feature $\mathbf{B}_\text{L}$ and the voxel feature $\mathbf{V}_\text{L}$ are derived as follows:
\begin{equation}
\begin{gathered}
    \mathbf{B}_\text{L}  = f_\text{point}(\mathbf P_n) \in
    \mathbb{R}^{N \times H \times W\times C_\text{L}},
    \\
    \mathbf{V}_\text{L} = f_\text{voxel}(\mathbf P_n) \in \mathbb{R}^{N \times H \times W \times L\times C_\text{L}},
\end{gathered}
\end{equation}
where $(H, W, L)$ and $C_\text{L}$ represent voxel size and feature channel of LiDAR.
Furthermore, following the common depth supervision methods in single vehicle \cite{BEVDepth,CaDNN}, we obtain a sparse depth map from the point clouds and use depth completion \cite{depth-complete} to supervise the depth predicted by the camera branch.

\subsubsection{Camera branch.} Given the images $\{\mathbf{I}_n\}_{n=1}^N$, we generate image features $\mathbf{F}_\text{I} \in \mathbb{R}^{N \times H \times W \times C_I}$ using a standard image encoder \cite{ResNet}. We then use a Depth Net \cite{LSS,CaDNN,BEVDepth} to predict the depth distribution interval $\mathbf{D} \in \mathbb{R}^{N \times H \times W \times L}$. Next, we compute the outer product between $\mathbf{F}_\text{I}$ and $\mathbf{D}$, and apply intrinsic and extrinsic parameters for the view transformation, resulting in the image voxel representation $\mathbf{V}_\text{I} \in \mathbb{R}^{N \times H \times W \times L \times C_\text{I}}$.

\subsection{Modality Fusion}
\begin{figure}[htbp]
    \centering
    \includegraphics[width=0.55\linewidth]{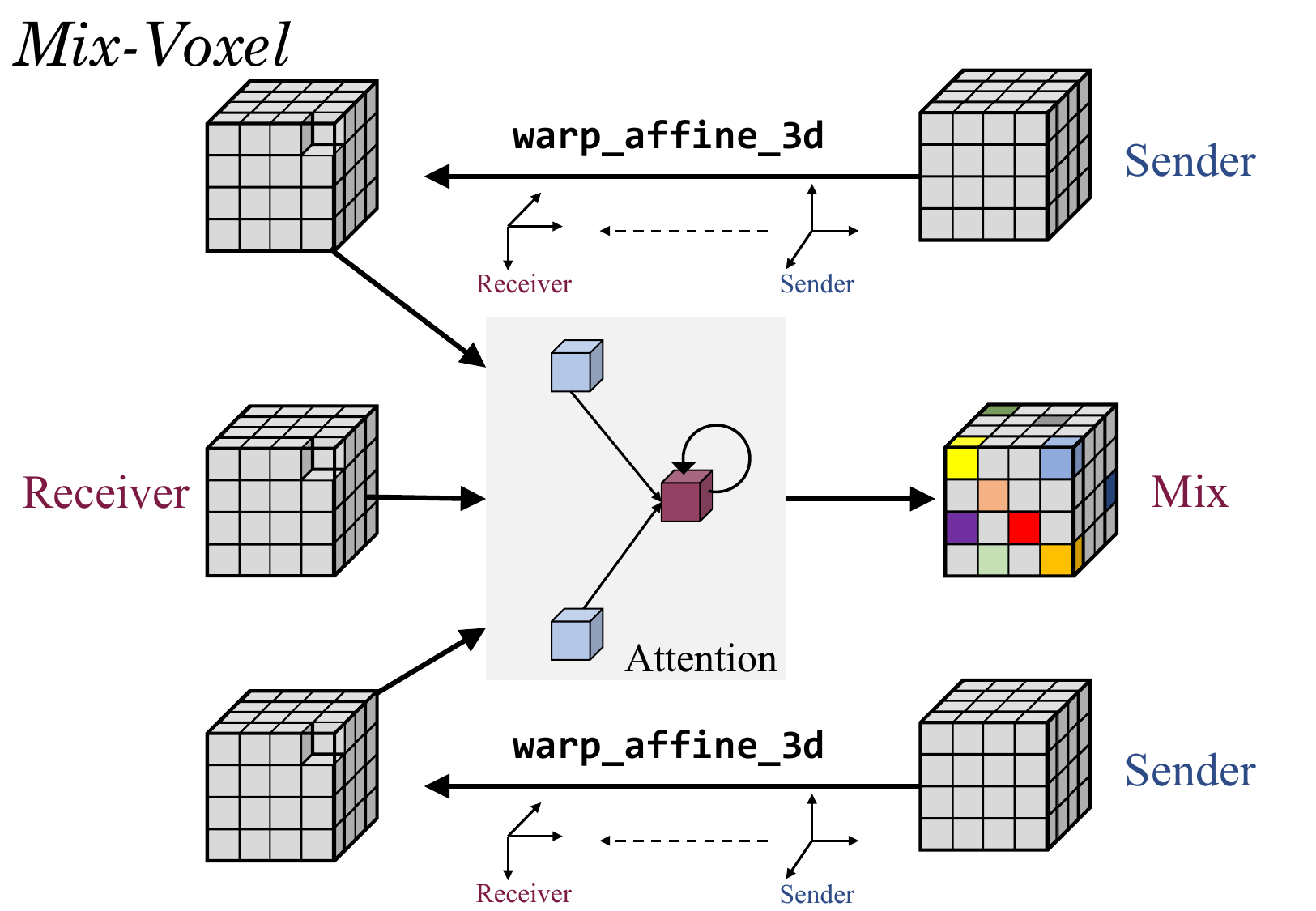}
    \caption{\textbf{Mix-Voxel module} constructs a local graph of voxels, utilizing self-attention mechanisms to facilitate information exchange. 
    }
    \label{mix-occ}
\end{figure}
To bridge distributional and spatial discrepancies between LiDAR and camera features, we propose the Occ-Guided Image Voxel Representation and the Heterogeneous Modality Fusion (HMF) module.
\subsubsection{Occ-Guided Image Voxel Representation.}
This component leverages LiDAR-based occupancy to enhance camera depth estimation.
As shown in Fig. \ref{mix-occ}, the Mix-Voxel module facilitates shared information by aligning all voxels to the ego-agent coordinate system and constructing a local graph. Self-attention on this graph models inter-agent voxel interactions to capture representative features:
\begin{equation}
\mathbf{V}_\text{mix} = f_\text{Mix-Voxel}(\mathbf{V}_\text{L}^\text{rc},\mathbf{V}_\text{L}^{\text{sd}_1},\cdots,\mathbf{V}_\text{L}^{\text{sd}_n}).
\end{equation}
Here, $\mathbf{V}_\text{L}^\text{rc}$ and $\mathbf{V}_\text{L}^\text{sd}$ represent the voxel associated with the ego agent and the neighboring agents, respectively. After that, $\mathbf{V}_\text{mix}$ is transmitted to $\Phi_\text{Occhead}$ to generate the occupancy probability of the 3D scene.
\begin{equation}
    \mathbf{O}_\text{L} = \Phi_\text{Occhead}(\mathbf{V}_\text{mix}), \in \mathbb{R}^{N \times H \times W \times L \times 1}.
\end{equation}
$\mathbf{O}_\text{L}$ shares the same resolution as $\mathbf{V}_\text{I}$. 
The camera voxel $\mathbf{V}_\text{I}$ is then multiplied by $\mathbf{O}_\text{L}$ to obtain the occ-guided image voxel representation:
\begin{equation}
    \mathbf{V}_\text{I}^\prime = \mathbf{V}_\text{I} \odot \mathbf{O}_\text{L}.
\end{equation}
Where $\odot$ denotes element-wise multiplication operation. To effectively aggregate geometric and depth information to $\mathbf{V}_\text{I}^\prime$, we add $\mathbf{V}_\text{mix}$ to $\mathbf{V}_\text{I}^\prime$. Finally, we obtain the camera BEV feature $\mathbf{B}_\text{I}$ through voxel collapse.

\subsubsection{Heterogeneous Modality Fusion.}
\begin{figure}[htbp]
    \centering
    \includegraphics[width=\linewidth]{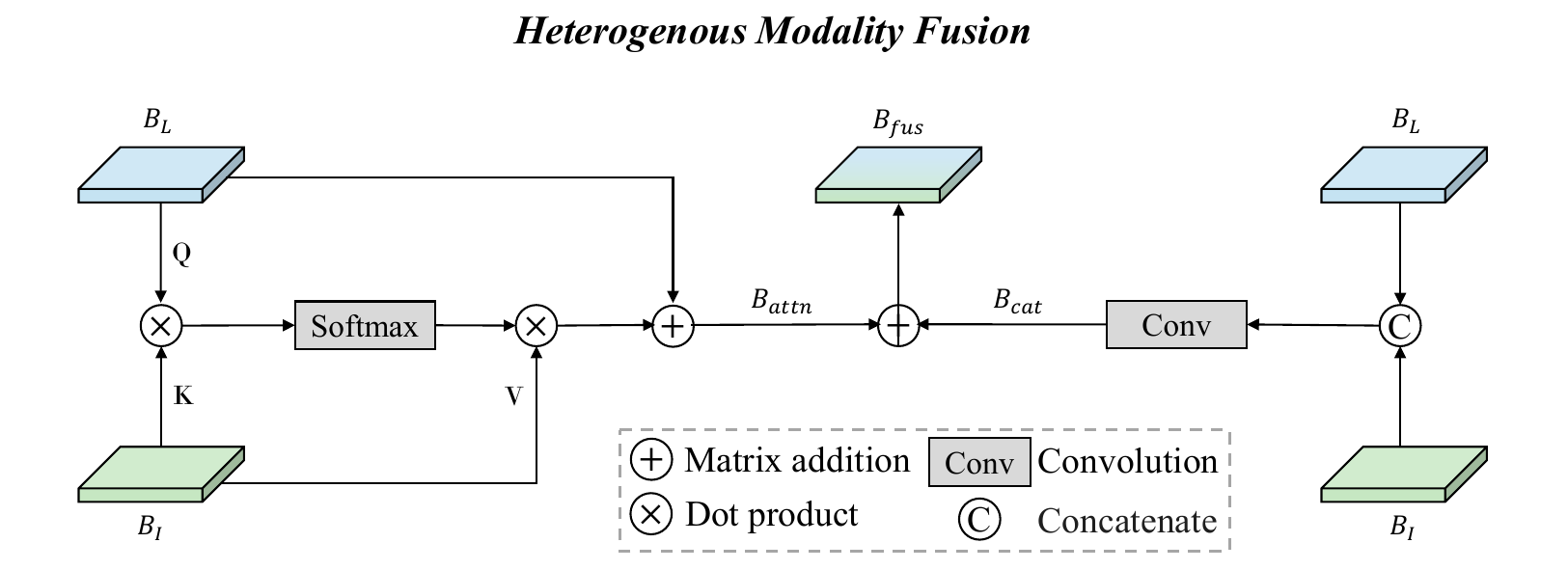}
    \caption{\textbf{Heterogeneous Modality Fusion module}. The module consists of two streams. The left part uses an attention-based method to establish the interaction between the two modalities, while the right part employs a basic operation (e.g., concatenation) followed by convolutional layers to fuse the modalities.}
    \label{HMF}
\end{figure}

Unlike conventional methods that simply concatenate BEV features , our HMF module (Fig. 4 ) effectively aggregates semantic and geometric information in a unified BEV space.
The detailed architecture is illustrated in Fig. \ref{HMF}. We use $1 \times 1$ convolutions to expand $\mathbf{B_\text{L}}$ and $\mathbf{B_\text{I}}$ to appropriate channels $C$, then concatenate the expanded BEV features to obtain feature $\mathbf{B_\text{cat}}$. Additionally, we employ attention mechanisms to facilitate interaction between the LiDAR and image BEV features. Since the LiDAR BEV feature is more reliable than the camera's, we define $\mathbf{B_\text{L}}$ as the query. Formally, it can be expressed as:
\begin{equation}
    \mathbf{B}_\text{attn} = \text{MLP}(\text{softmax}(\frac{\mathbf{B_\text{L}} \cdot \mathbf{B_\text{I}}}{\sqrt{C}})\cdot \mathbf{B_\text{I}}) + \mathbf{B_\text{L}}
\end{equation}
The final fused BEV features can be presented as:
\begin{equation}
    \mathbf{B_\text{fus}} = \mathbf{B_\text{cat}} + \mathbf{B}_\text{attn}, \in \mathbb{R}^{N \times H \times W \times C}  .
\end{equation}

\subsection{Collaboration}
\begin{figure}[htbp]
    \centering
    \includegraphics[width=\linewidth]{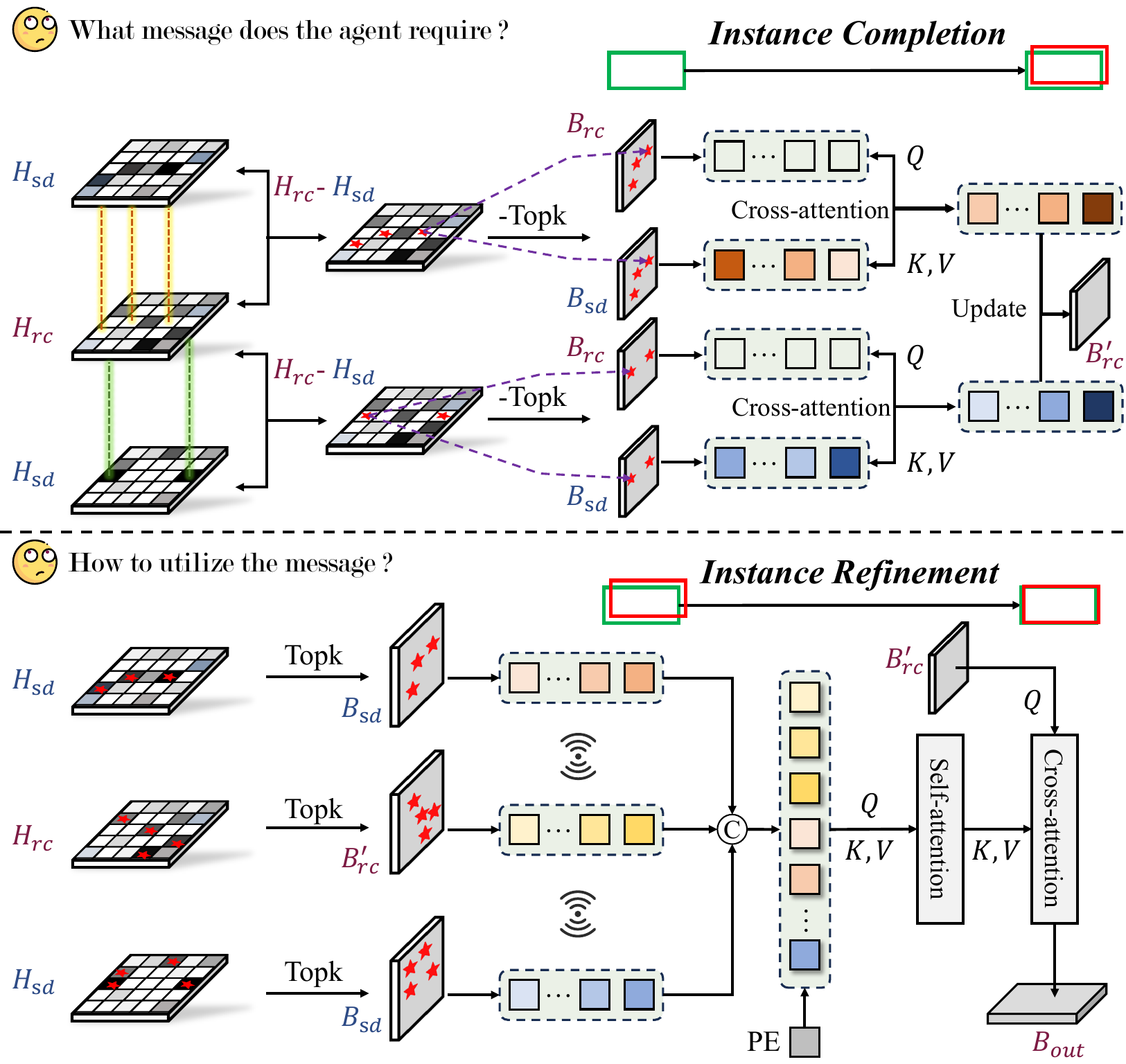}
    \caption{\textbf{Instance-level communication}. The Instance Completion (IC) module prioritizes critical regions by analyzing cross-agent heatmap discrepancies, performing instance completion via cross-attention. The IR module first selects agent-specific instances from heatmaps, then refines them via self-attention. Finally, it aggregates instance-to-scene context by cross-attending BEV features (query) to instance representations (key/value).}
    \label{Instance collaboration}
\end{figure}

After obtaining $\mathbf{B_\text{fus}}$, we focus on how to transmit messages effectively and efficiently. Existing collaborative perception paradigms often suffer from prohibitive bandwidth costs due to dense BEV feature transmission. TTransIFF \cite{TransIFF} introduces sparse instances to reduce communication bandwidth, but it does not achieve optimal performance and is not suitable for multimodal scenarios. In this work, we propose a novel heatmap-driven instance-level communication strategy, comprising two modules, Instance Completion and Instance Refinement, to balance performance and bandwidth, as illustrated in Figure \ref{Instance collaboration}.

\subsubsection{Instance Completion.}
The core objective of this module is to determine what complementary perceptual message the ego agent (receiver) requires from collaborating agents (sender). Specifically, for each agent pair (receiver rc and one sender sd), we generate heatmaps $\mathbf{H}_\text{rc}$ and $\mathbf{H}_\text{sd} \in \mathbb{R}^{H \times W \times 1}$ through a lightweight CNN $\Phi_\text{hm}$. The target discrepancy heatmap is computed as:
\begin{equation}
    \mathbf{H}_\text{tg} = \mathbf{H}_\text{rc} - \mathbf{H}_\text{sd} \in \mathbb{R}^{(N-1) \times H \times W \times 1}.
\end{equation}
We then identify the $\mathbf{K}_\text{IC}$ spatial positions with minimal values in $\mathbf{H}_\text{tg}$, indicating regions where the receiver's perception is least confident compared to the sender. Then, for each selected position $(h,w)$, we extract receiver's BEV feature $\mathbf{F}_\text{rc}^{(h,w)} \in \mathbb{R}^C$ as query $\mathbf{Q}$ and sender's BEV feature $\mathbf{F}_\text{sd}^{(h,w)}$ as key/value ($\mathbf{K},\mathbf{V}$). These features undergo cross-attention:
\begin{equation}
\mathbf{F}_{rc}^{\text{updated}} = \text{Softmax}\left(\frac{\mathbf{Q}\mathbf{K}^T}{\sqrt{C}}\right)\mathbf{V}.
\end{equation}
When multiple senders update the same spatial position in $\mathbf{B}_\text{rc}$,  we employ element-wise summation for feature fusion:
\begin{equation}
 \mathbf{B}_{rc}^{(h,w)} \leftarrow \sum_{s=1}^{N_{\text{senders}}} \mathbf{F}_{rc,s}^{\text{updated}}
\end{equation}
The insight behind this module is that the key aspect of collaborative perception lies in the transmission of complementary information. This complementary information is reflected in the heatmap as areas where the ego has low confidence in the presence of an object, while the sender exhibits high confidence. As a result, subtracting the sender's heatmap from the ego's heatmap highlights the areas of interest for the ego. Through the heatmap-driven approach, the most relevant and critical regions can be effectively identified and prioritized for further processing.

\subsubsection{Instance Refinement.}
Following instance completion, this module aims to holistically refine the potential instances through attention mechanisms. First, we extract Top-k critical region $(h,w)$ from heatmap $\mathbf{H}_n \in \mathbb{R}^{H \times W}$ for each agent. Then we retrieve instance features $\mathbf{F}_n \in \mathbb{R}^{\mathbf{K}_\text{IR} \times C}$ from corresponding BEV features $\mathbf{B}_n$, at position $(h,w)$. We concatenate all instances:
\begin{equation}
    \mathbf{F}_\text{all} = [\mathbf{F_\text{rc}};\mathbf{F_{\text{sd}_1}};\cdots;\mathbf{F_{\text{sd}_N}}] \in \mathbb{R}^{N\cdot \mathbf{K}_\text{IR} \times C}.
\end{equation}
To enable information exchange from a data-driven perspective, we employ self-attention for inter-instance communication:
\begin{equation}
    \mathbf{F}_\text{all}^\text{updated} = \text{SelfAttn}(\mathbf{F}_\text{all},\mathbf{F}_\text{all},\mathbf{F}_\text{all}),
\end{equation}
where $\mathbf{F}_\text{all}$ is addition with positional encoding $E_\text{pos}(h,w)$ maintaining spatial relationships. Finally, cross-attention enables each BEV grid feature to acquire valuable information from potentially relevant instances:
\begin{equation}
    \mathbf{B}_\text{out} = \text{CrossAttn}(Q_\text{scence}, K_\text{ins}, V_\text{ins}),
\end{equation}
\begin{equation}
    Q_\text{scence} = \mathbf{B}_\text{rc}^\prime \in \mathbb{R}^{H\times W \times C},
\end{equation}
\begin{equation}
    K_\text{ins},V_\text{ins} = \mathbf{F}_\text{all}^\text{updated}.
\end{equation}
This module enables adaptive message passing between instances via self-attention and context-aware scene reconstruction through instance-to-BEV cross-attention with minimal communication bandwidth.

To capture instances and scenes at varying granularities and enhance the information exchange between agents, we downsample the initial feature map to generate feature maps at different scales for use in the collaboration stage. Finally, the feature maps from all scales are fused to produce an output feature map that matches the dimensions of the original input.

\subsection{Loss Function}
Finally, the training loss of the model is simply the sum of the regression loss $L_\text{reg}$, classification loss $L_\text{cls}$, depth estimation loss $L_\text{depth}$, direction loss $L_\text{dir}$ and heatmap loss $L_\text{hm}$:
\begin{multline}
    L_\text{total} = \lambda_\text{reg} L_\text{reg} + \lambda_\text{cls} L_\text{cls} + \lambda_\text{depth} L_\text{depth} \\
    + \lambda_\text{dir} L_\text{dir} + \lambda_\text{hm} L_\text{hm}.
\end{multline}
Where the $\lambda$ are the weighting factors of the different losses used in the optimization process.

\section{Experimental Results}

\begin{table}[htbp]
\caption{Comparison of mainstream works on the OPV2V and DAIR-V2X dataset. The best performance is highlighted in \textbf{bold}. $\text{Ours}^\dagger$ indicates the model without the Mix-Voxel module.}
\resizebox{\linewidth}{!}{
\begin{tabular}{c|ccc|ccc|c}
\Xhline{3\arrayrulewidth}
                         & \multicolumn{3}{c|}{OPV2V} & \multicolumn{3}{c|}{DAIR-V2X} & \multicolumn{1}{c}{Bandwidth} \\ \cline{2-8}
\multirow{-2}{*}{Method} & \multicolumn{1}{c|}{AP30} & \multicolumn{1}{c|}{AP50} & \multicolumn{1}{c|}{AP70} & \multicolumn{1}{c|}{AP30} & \multicolumn{1}{c|}{AP50} & \multicolumn{1}{c|}{AP70} & \multicolumn{1}{c}{Comm ($\log_2$)} \\ \hline
No Coll                                          & 83.63 & 63.74 & 58.32 & 69.99 & 65.02 & 53.82 & 0.00    \\
Fcooper                                          & 93.88 & 89.03 & 74.28 & 76.61 & 69.29 & 51.37 & 24.00   \\
Attn                                             & 88.08 & 86.30 & 75.32 & 68.78 & 63.16 & 49.17 & 24.00   \\
V2VNet                                           & 93.56 & 93.13 & 89.00 & 77.36 & 72.22 & 52.95 & 25.43   \\
V2X-VIT                                          & 95.09 & 93.66 & 86.06 & 77.29 & 71.87 & 55.46 & 24.00   \\
CoBEVT                                           & 94.54 & 93.03 & 84.64 & 77.86 & 71.70 & 55.85 & 24.00   \\
Where2comm                                       & 88.36 & 86.77 & 76.34 & 68.18 & 63.16 & 51.04 & 24.00   \\
BM2CP                                            & 93.34 & 93.04 & 88.94 & \textbf{77.91} & 72.37 & 56.18 & 23.18   \\ \hline

$\text{Ours}^\dagger$                                             & 93.59  & 92.27 & 83.96 & 76.82 & 71.81 & 56.76 & 18.97
\\

Ours                                             & \textbf{95.29} & \textbf{94.71} & \textbf{89.16} & 75.01 & \textbf{73.01} & \textbf{58.37} & 20.16

\\ \Xhline{3\arrayrulewidth}
\end{tabular}}

\vspace{-8pt}
\label{Performance}
\end{table}

\subsection{Performance}
As evidenced in Table \ref{Performance}, our EIMC framework establishes new state-of-the-art detection performance across both benchmark datasets. On the OPV2V dataset, EIMC achieves superior results with 95.29\% AP30, 94.71\% AP50, and 89.16\% AP70, outperforming existing approaches by margins of +0.20\%, +1.58\%, and +0.16\%, respectively, in these metrics.
The framework exhibits particularly strong robustness on the DAIR-V2X dataset, attaining 58.37\% AP70, a 2.19\% absolute improvement over the previous best method (BM2CP: 56.18\%). This significant advancement at the strictest IoU threshold (0.7) demonstrates EIMC’s ability to ensure reliable perception under real-world challenges.
As shown in the table, EIMC effectively balances communication overhead and performance.
Following the \cite{where2comm}, the communication volume is computed as follows:
\begin{equation}
    Comm(B) = \log_2(H \times W \times C \times \text{float32} / 8).
\end{equation}
\begin{figure}[htbp]
    \centering
    \includegraphics[width=\linewidth]{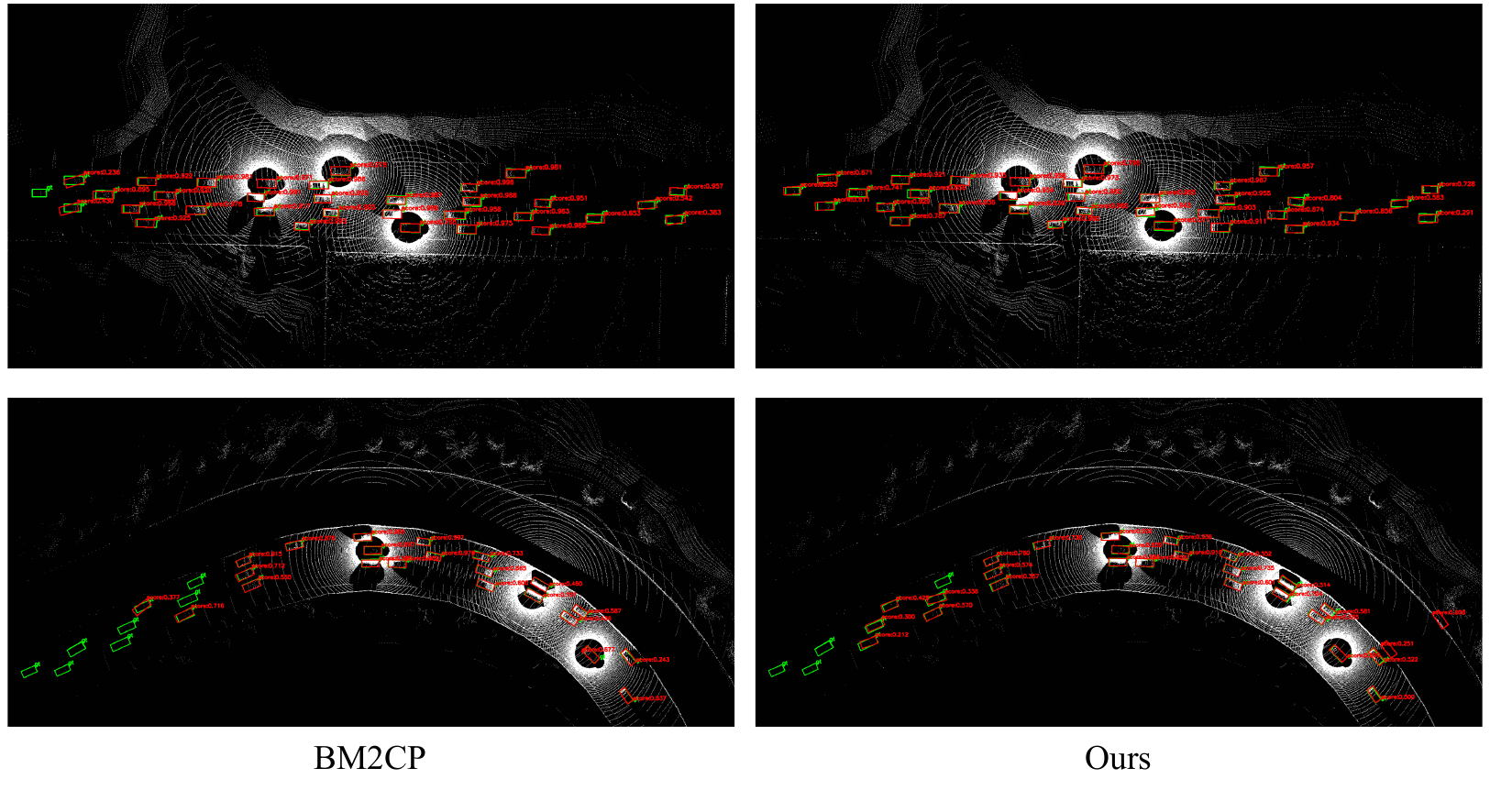}
    \caption{Visualization of predictions from BM2CP and EIMC on OPV2V dataset.}
    \label{vis_res}

    \vspace{-12pt}
\end{figure}
Fig. \ref{vis_res} shows the comparison with BM2CP and EIMC. EIMC achieves more complete and accurate detection. To evaluate the effectiveness of our collaborative perception method, we compared the performance of different LiDAR-only methods, as shown in Table \ref{LiDAR-only}. Our method achieves competitive results, with AP50 reaching 68.46\% and AP70 reaching 54.02\%.
Note that the performance of LiDAR-only methods may outperform that of multimodal methods due to the following reasons: 1) DAIR-V2X is a real-world dataset, which introduces sensor alignment noise; 2) Unlike the OPV2V dataset, DAIR-V2X only provides front-view cameras for vehicles and roadside cameras, limiting the camera branch’s ability to obtain surround-view BEV features.

\begin{table}[h]
\vspace{-6pt}
  \centering
  \caption{LiDAR-only performance comparison of different methods on DAIR-V2X dataset.}
  \label{LiDAR-only}
  \begin{tabular*}{\linewidth}{@{\extracolsep{\fill}} l S[table-format=2.2] S[table-format=2.2] S[table-format=2.2]}
    \toprule
    \textbf{Method}         & \textbf{AP30} & \textbf{AP50} & \textbf{AP70} \\
    \midrule
    No Coll                 & 65.26 & 59.88 & 48.52 \\
    Fcooper                 & 73.10 & 65.36 & 44.52 \\
    Attn                    & 69.85 & 64.33 & 51.12 \\
    V2VNet                  & 66.92 & 53.60 & 49.83 \\
    V2X-VIT                 & 67.14 & 63.30 & 49.34 \\
    CoBEVT                  & \textbf{75.07} & 67.70 & 47.09 \\
    Where2comm              & 69.48 & 64.46 & 51.77 \\
    Ours                    & 73.22 & \textbf{68.47} & \textbf{54.02} \\
    \bottomrule
  \end{tabular*}
  \vspace{-6pt}
\end{table}

\subsection{Ablation}
\begin{table}[htbp]
\centering
\caption{Ablation Studies of Components on the DAIR-V2X Dataset.}
\resizebox{\linewidth}{!}{%
\begin{tabular}{cccc|rrr|r}
\Xhline{2\arrayrulewidth} 
\textbf{MV} & \textbf{IC} & \textbf{IR} & \textbf{MS} & \textbf{AP30} & \textbf{AP50} & \textbf{AP70} & \textbf{\#Params (M)} \\
\hline     		
- & - & - & -  & 70.25 & 67.73 & 50.91 & $\sim$25.9 \\
\checkmark & - & - & - & 71.02 & 68.39 & 51.44 & $\sim$26.1 \\
\checkmark & \checkmark & - & - & 74.12 & 69.34 & 54.78 & $\sim$26.8 \\
\checkmark & - & \checkmark & - & 71.77 & 68.24 & 52.09 & $\sim$28.0 \\
- & \checkmark & \checkmark & \checkmark & \textbf{76.82} & 71.81 & 56.76 & $\sim$42.0 \\
\checkmark & \checkmark & \checkmark & - & 74.22 & 70.54 & 56.28 & $\sim$30.9 \\
\checkmark & \checkmark & \checkmark & \checkmark & 75.01 & \textbf{73.01} & \textbf{58.37} & $\sim$42.0 \\
\Xhline{2\arrayrulewidth} 
\end{tabular}%
}

\label{components ablations}
\end{table}
\textbf{Components analysis.} Our ablation studies on DAIR-V2X, as shown in Table \ref{components ablations}, demonstrate critical insights: the IC module drives the most substantial performance gain, boosting AP70 by +3.34\% ($51.44\% \to 54.78\%$) through targeted cross-agent feature retrieval. The MV module significantly enhances perception accuracy. Ultimately, the full framework with MS integration achieves optimal results (58.37\% AP70, 42M parameters), highlighting the synergistic benefits of our design. 

\begin{table}[h]
\centering
\caption{Ablation study of Top-k selection strategies on DAIR-V2X Dataset.}
\resizebox{\linewidth}{!}{ 
\small
\begin{tabular}{@{}cS[table-format=2.2]S[table-format=2.2]S[table-format=2.2]@{\hspace{1em}}cS[table-format=2.2]S[table-format=2.2]S[table-format=2.2]@{}}
\toprule
\multicolumn{4}{c}{\textbf{Instance Completion}} & \multicolumn{4}{c}{\textbf{Instance Refinement}} \\
\cmidrule(r{0.5em}){1-4} \cmidrule(l{0.5em}){5-8}
{$\mathbf{K}_\text{IC}$} & {AP30} & {AP50} & {AP70} & {$\mathbf{K}_\text{IR}$} & {AP30} & {AP50} & {AP70} \\
\midrule
10    & 73.46 & 68.69 & 53.51 & 200/100/50 & \textbf{75.93} & 71.19 & 56.52 \\
15    & 73.28 & 68.16 & 52.68 & 150/100/50 & 75.09 & 70.82 & 56.72 \\
20    & 75.01 & \textbf{73.01} & \textbf{58.37} & 100/50/25 & 75.01 & \textbf{73.01} & \textbf{58.37} \\
25    & \textbf{77.78} & 72.15 & 54.96 & 100/100/100 & 74.34 & 69.84 & 54.70 \\
30    & 75.71 & 70.68 & 54.57 & 50/50/50 & 72.39 & 68.96 & 54.12 \\
\bottomrule
\end{tabular}
}

\label{tab:topk}
\end{table}

\subsubsection{Analysis on Parameters.} As shown in Table \ref{tab:topk}, we perform the analysis on the number of Top-k selection. It can be observed that if $\mathbf{K}_\text{IC}$ is too small, it may fail to capture all potential key regions adequately, whereas an excessively large $\mathbf{K}_\text{IC}$ may reduce the accuracy of the bounding box. Moreover, dynamically adjusting the number of $\mathbf{K}_\text{IR}$ according to scale variations can improve detection performance.
\begin{figure}[ht]
    \centering
    \includegraphics[width=0.4\textwidth]{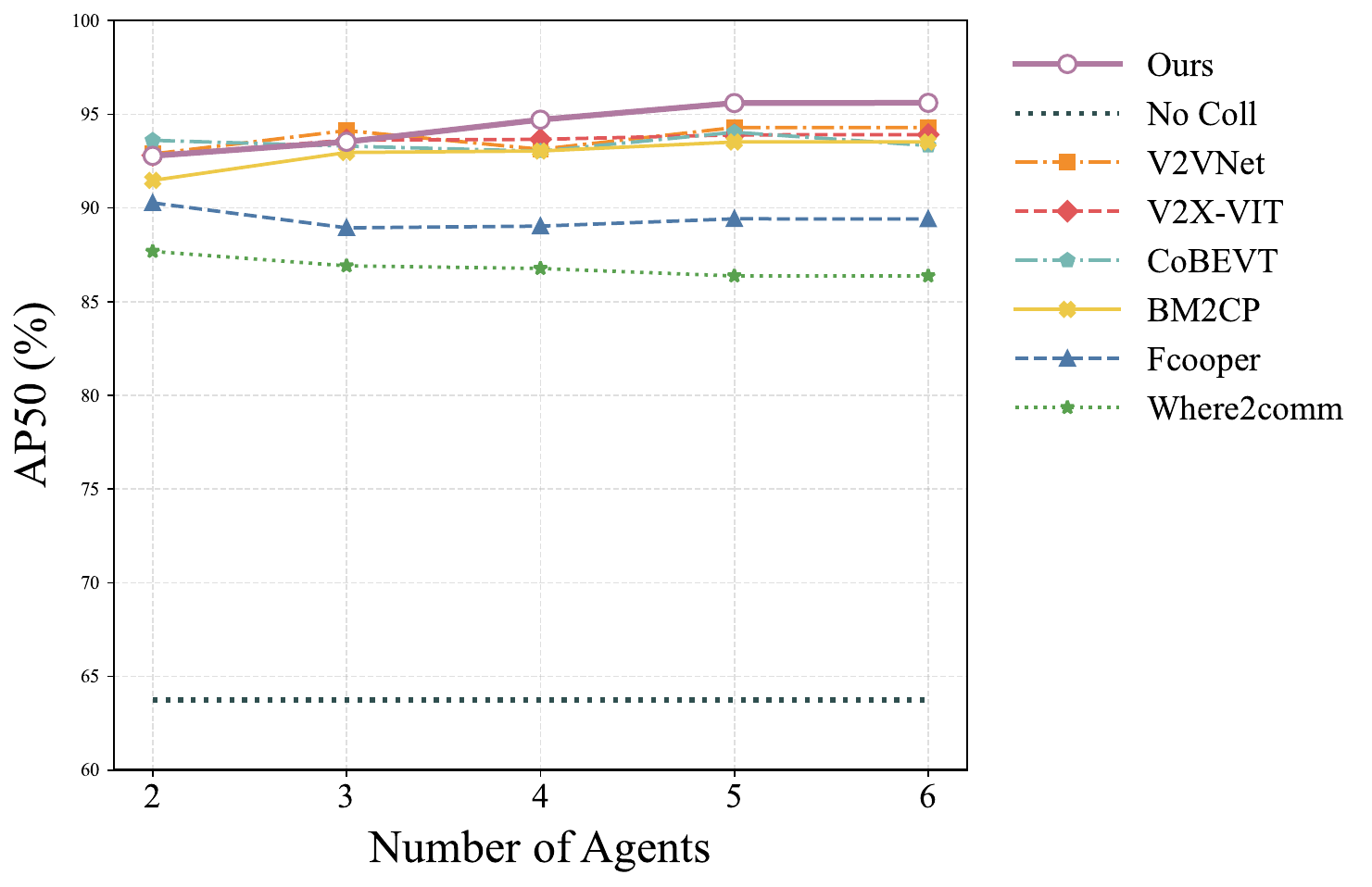}
    \caption{Average Precision at IoU=0.5 with respect to agent number. Note that all methods were trained using the default agent setting of 4, and the results were obtained through inference with varying agent numbers. With the exception of CoBEVT, which, due to the inherent limitations of the method, requires retraining based on the number of agents.}
    \label{number of agents}
\end{figure}
We analyze how detection accuracy (AP) varies with the number of agents (Fig. \ref{number of agents}). While most methods initially improve with more agents, gains saturate beyond 3–4 agents due to scene coverage limits. Our framework effectively enhances multi-agent cooperation, achieving clear accuracy improvements as agent numbers increase within environmental constraints.

\begin{figure}[htbp]
    \centering
    \includegraphics[width=\linewidth]{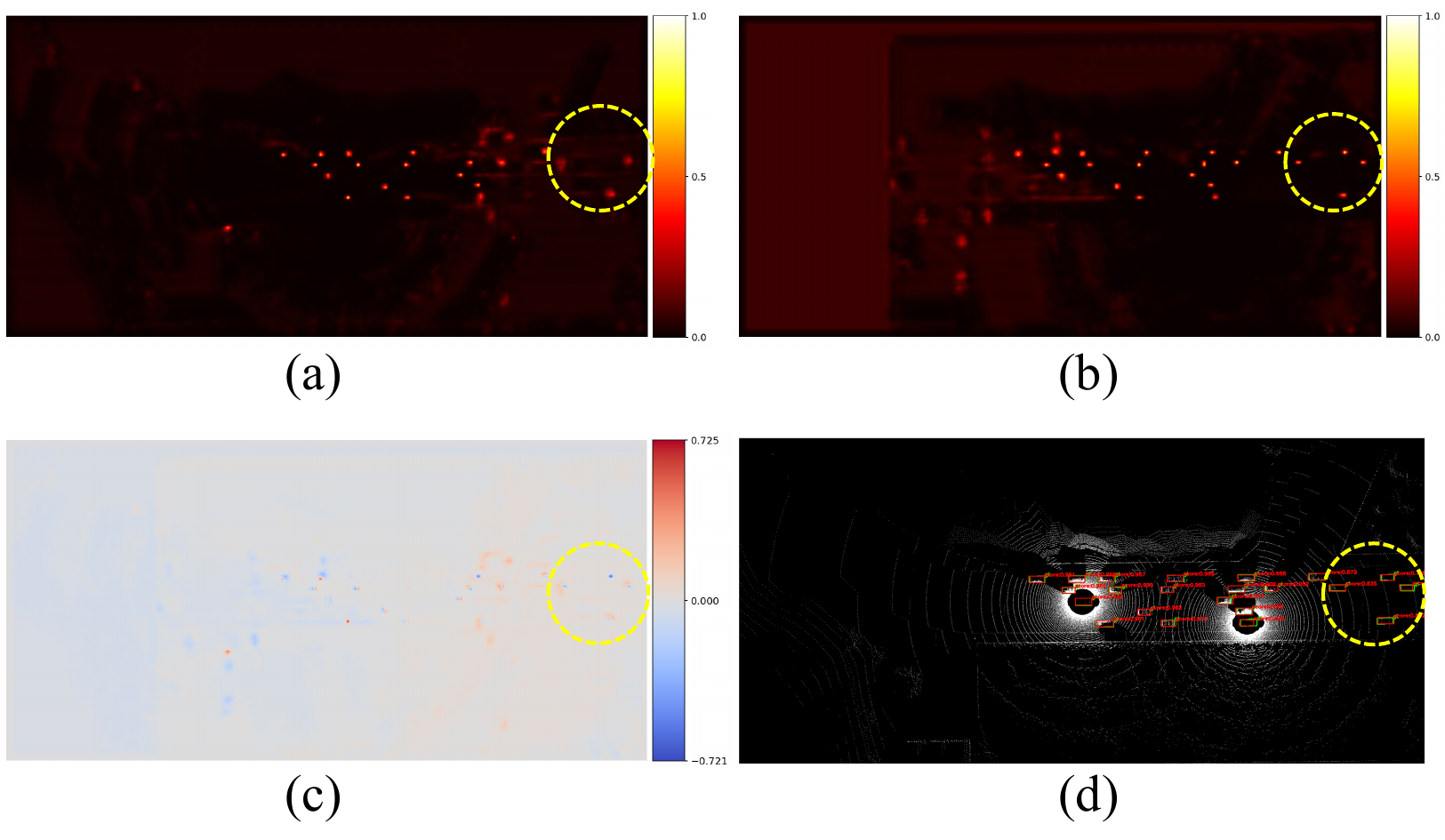}
    \caption{\textbf{Visualization of heatmap and result.} (a) Ego agent heatmap; (b) Communication agent heatmap; (c) Target heatmap; (d) Output. The area within the yellow circle highlights regions of interest.}
    \label{heatmap}
\end{figure}

\begin{table*}[htbp]
\centering
\footnotesize
\caption{Robustness analysis under pose noise on OPV2V and DAIR-V2X. The pose noise follows a Gaussian distribution.}
\setlength{\tabcolsep}{3.2pt}
\begin{tabularx}{\textwidth}{@{}l*{16}{Y}@{}}
\toprule
\multirow{2}{*}{Method} & 
\multicolumn{8}{c}{\textbf{AP50 under Noise Level ($\sigma_p$/$\sigma_r$)}} &
\multicolumn{8}{c}{\textbf{AP70 under Noise Level ($\sigma_p$/$\sigma_r$)}} \\
\cmidrule(lr){2-9} \cmidrule(l){10-17}
& \multicolumn{4}{c}{OPV2V} & \multicolumn{4}{c}{DAIR-V2X} &
\multicolumn{4}{c}{OPV2V} & \multicolumn{4}{c}{DAIR-V2X} \\
\cmidrule(lr){2-5} \cmidrule(lr){6-9} \cmidrule(lr){10-13} \cmidrule(l){14-17}
& 0/0 & 0.2/0.2 & 0.4/0.4 & 0.6/0.6 & 0/0 & 0.2/0.2 & 0.4/0.4 & 0.6/0.6 &
0/0 & 0.2/0.2 & 0.4/0.4 & 0.6/0.6 & 0/0 & 0.2/0.2 & 0.4/0.4 & 0.6/0.6 \\
\midrule
No Coll          &  63.74   &  -      &   -      &  -       &  65.02   &    -     &   -      &    -     &  58.32   &    -     &     -    &   -      & 53.82    &   -      &   -      &    -     \\
Fcooper          &  89.03   &  80.77       &   66.53      &  59.30       &  69.29   &   67.43      &  64.18       &   62.25      &  74.28   &  62.62      &  51.56       &  47.29       & 51.37    &   50.18      &  48.46    &  47.56       \\
Attn             &  88.08   &  86.31       &   80.57      &  77.98       &  63.16   &    57.32     &  54.19       &   52.06      &  75.32   &   73.33      &  69.09      &  60.23       & 49.17    &   45.77      &    43.17     &  42.75       \\
V2VNet           &  93.13   &  91.04       &   74.16      &  61.45      &  72.32   &    68.76     &   62.21     & 58.48      &  89.00       &  67.33     &   37.96      &  28.55       & 52.95    &  46.97     &   43.00      &  40.85      \\
V2X-VIT          &  93.66   &  91.87       &   86.29      &  80.33       &  71.87   &   69.13      &  64.86       &   61.95      &  86.06   &   77.63      &  65.64       &  60.04       &  55.46    &  52.47       &  50.44       &  49.17       \\
CoBEVT           &  93.03   & 91.14        &  84.29       &  76.27    &  71.70   &  69.26       &   64.75      &  62.46     & 84.64       &   77.69      &  65.28    &  57.41       & 55.85    &   53.44      &  51.12       &  50.14       \\
Where2comm       &  86.77   & 80.34        &  76.57       &  72.19       &  63.16   &  63.12       &   62.47      &   60.14      &  76.34   &  68.73       &  60.32       &  54.77      & 51.04    &  51.04       &  51.03       &  51.02       \\
BM2CP            &  93.04   &   91.34      &  81.55       &   73.54      &  72.17   &  \textbf{70.31}       &  \textbf{65.07}       &   61.78      &  88.94   & 75.35        &  57.43       &  49.53       & 56.18    &   53.24      & 49.97        &    48.65     \\
Ours             &  \textbf{94.71}   &   \textbf{92.44}      &  \textbf{89.78}       &   \textbf{87.87}      &  \textbf{73.01}   &   68.23      &    64.86     & \textbf{63.82}        &  \textbf{89.16}   &  \textbf{80.60}       &  \textbf{76.33}       &   \textbf{73.98}      & \textbf{58.37}    &  \textbf{55.18}       &   \textbf{52.91}      &  \textbf{52.21}       \\
\bottomrule
\end{tabularx}

\label{pose_error}

\end{table*}

\subsubsection{Analysis of Heatmap.} The Instance Completion module mainly addresses the occlusion problem that cannot be resolved by a single agent. As shown in Fig. \ref{heatmap}, (a) depicts the ego agent's heatmap, (b) shows the heatmap of a neighboring agent, and (c) represents the target heatmap. It is clearly observed that, within the yellow region, the blue instances are highlighted. These instances are occluded and have low confidence from the ego's perspective. Through the Instance Completion module, these instances can ultimately be reconstructed and detected.

\subsubsection{Robustness to localization noise.} We also evaluate the robustness to localization noise following the setting in \cite{where2comm}. The results are shown in Table \ref{pose_error}. For each noise level, we compare the AP performance of EIMC at IOU thresholds of 0.5 and 0.7 with several other methods. As shown in the table, most existing methods exhibit a noticeable decrease in performance as noise levels increase at an IOU of 0.7. In contrast, EIMC demonstrates remarkable robustness, maintaining high performance even under severe noise conditions. Furthermore, under the most extreme noise conditions, in the OPV2V dataset, EIMC outperforms the best-performing method by 9.3\% at an IOU of 0.5 and by 22.8\% at an IOU of 0.7. In the DAIR-V2X dataset, the performance improvements are 2.1\% and 2.3\%, respectively.

\subsubsection{HMF module.}
We performed an ablation study on the HMF module. As shown in Table~\ref{HMF_ab}, relying solely on the attention mechanism is insufficient for fully aligning heterogeneous modalities, whereas integrating the concatenation operation yields optimal performance.
\begin{table}[tbp]
\centering
\caption{Ablation studies of HMF module.}
\label{HMF_ab}

\setlength{\tabcolsep}{6pt}
\renewcommand{\arraystretch}{1.08}

\begin{tabular*}{\columnwidth}{@{\extracolsep{\fill}} l
  S[table-format=2.2] S[table-format=2.2] S[table-format=2.2] @{}}
\toprule
\textbf{Strategy} & \textbf{AP30} & \textbf{AP50} & \textbf{AP70} \\
\midrule
Attn        & 74.18 & 68.80 & 53.94 \\
Attn+Concat & 75.01 & \textbf{73.01} & \textbf{58.37} \\
Attn+Add    & \textbf{75.23} & 70.94 & 58.06 \\
\bottomrule
\end{tabular*}
\vspace{-16pt}
\end{table}

\subsubsection{Mix-Voxel module}
\begin{table}[t]
\centering
\caption{Ablation studies of MV module.}
\label{mix-voxel-compression}

\setlength{\tabcolsep}{6pt}
\renewcommand{\arraystretch}{1.08}

\begin{tabular*}{\columnwidth}{@{\extracolsep{\fill}} l
  S[table-format=2.2] S[table-format=2.2] S[table-format=2.2] S[table-format=2.2] @{}}
\toprule
Method & {AP30} & {AP50} & {AP70} & {Bandwidth($\log_2$)}\\
\midrule
M1 & 75.01 & 73.01 & 58.37 & 20.16 \\
M2 & 74.22 & 72.35 & 56.96 & 19.67 \\
M3 & \textbf{76.24} & \textbf{73.55} & \textbf{59.17} & 22.46 \\
\bottomrule
\end{tabular*}

\vspace{-10pt}
\end{table}

Table~\ref{mix-voxel-compression} details our ablation study on the Mix-Voxel module's compression. We evaluated three strategies: an aggressive 4$\times$ downsampling (M2), a mild 2$\times$ downsampling (M3), and an asymmetric approach (M1). The results show that M3 achieves the highest accuracy but with substantial communication overhead, while M2 harms performance. Consequently, we select M1, which strikes the best balance between performance and efficiency.

\section{CONCLUSIONS}

By distilling sparse key instances from dense BEV features and leveraging dual completion and refinement mechanisms, the EIMC framework achieves an effective balance between communication efficiency and perception performance, while also demonstrating robust resilience to noise. In future work, we will evaluate the effectiveness of our method across additional datasets.

{
    \small
    \bibliographystyle{IEEEtran}
    \bibliography{main}
}

\end{document}